\documentclass[runningheads]{llncs}

\usepackage{acronym}
\usepackage{listings}
\usepackage{amsfonts}       
\usepackage{amsmath}		
\usepackage{amssymb}		
\usepackage{booktabs} 		
\usepackage[english]{babel} %
\usepackage{caption}        
\usepackage{subcaption}
\usepackage{comment}		
\usepackage[inline]{enumitem} 
\usepackage{float}			%
\usepackage[utf8]{inputenc}
\usepackage{graphicx}
\usepackage{hyperref}
\usepackage{bbm}
\usepackage{listings}		

\usepackage{booktabs}
\usepackage{multirow}
\usepackage{multicol}

\usepackage{mathtools}
\usepackage{microtype}      

\usepackage{nicefrac}       
\usepackage[binary-units=true]{siunitx}        
\usepackage{subcaption}     
\usepackage{tikz} 
\usepackage{todonotes}      
\usepackage{xcolor}			
\usepackage{xspace}

\usepackage{tabularx}
\usepackage{booktabs}
\usepackage{pifont}

\xdefinecolor{tugreen}{RGB}{132, 184, 24} 
\newcolumntype{Y}{>{\centering\arraybackslash}X}

\newacro{CNN}[CNN]{Convolutional Neural Network} 

\newacro{ML}[ML]{Machine Learning}

\newacro{pgm}[PGM]{probabilistic graphical model}

\newcommand{\pxpy}{\texttt{pxpy}\@\xspace}

\newacro{mrf}[MRF]{Markov random field}

\newacro{trwbp}[TRWBP]{Tree-Reweighted Belief Propagation}

\newacro{spn}[SPN]{Sum-Product Network}

\newacro{pac}[PAC]{Probably Approximately Correct}

\newacro{strf}[STRF]{Spatio-Temporal Random Field}

\newacro{lbp}[LBP]{loopy belief propagation}

\newacro{jt}[JT]{junction tree}

\newacro{pgd}[PGD]{projected gradient descent}

\newacro{fgsm}[FGSM]{fast gradient sign method}

\newacro{flops}[FLOPS]{floating point operations}
\newcommand{\flops}{\ac{flops}\@\xspace}

\newacro{dnn}[DNN]{deep neural network}

\newacro{alexnet}[AlexNet]{\texttt{AlexNet}}
\newcommand{\alexnet}{\ac{alexnet}\@\xspace}

\newacro{resneteighteen}[ResNet-18]{\texttt{ResNet-18}}
\newcommand{\resneteighteen}{\ac{resneteighteen}\@\xspace}

\newacro{mobilenetlarge}[MobileNetV3\_Large]{\texttt{MobileNetV3\_Large}}
\newcommand{\mobilenetlarge}{\ac{mobilenetlarge}\@\xspace}

\acused{alexnet}
\acused{resneteighteen}
\acused{mobilenetlarge}
\newcommand{\vgg}{\texttt{VGG11}\@\xspace}
\newacro{gpu}[GPU]{graphics processing unit}
\newcommand{\gpu}{GPU\@\xspace}

\newacro{cpu}[cpu]{central processing unit}
\newcommand{\cpu}{CPU\@\xspace}

\makeatletter
\newcommand\inlinecomment{\begingroup\@makeother\#\@inlinecomment}
\newcommand\@inlinecomment[2]{%
    \todo[%
        inline,%
        color=#1,%
    ]{#1:  #2}%
    \endgroup%
}%
\makeatother

\colorlet{DH}{blue!10}    
\colorlet{MJ}{green!10}   
\colorlet{KM}{red!20}     
\colorlet{LH}{blue!20}    
\colorlet{RF}{green!20}   
\colorlet{SM}{orange!20}
\colorlet{AP}{purple!20}
\colorlet{HK}{red!20}
\xdefinecolor{tugreen}{RGB}{132, 184, 24} 
\xdefinecolor{tudarkgreen}{RGB}{100, 150, 0} 
\xdefinecolor{tulightgreen}{RGB}{224, 234, 204} 
\xdefinecolor{tugray}{RGB}{207, 208, 210} 
\xdefinecolor{tugraybg}{RGB}{178, 179, 204} 
\xdefinecolor{tuorange}{RGB}{227, 105, 19} 
\xdefinecolor{tuyellow}{RGB}{242, 189, 0} 
\xdefinecolor{tucitron}{RGB}{249, 219, 0} 
\xdefinecolor{tublue}{RGB}{46, 134, 171} 
\xdefinecolor{tudarkblue}{RGB}{30, 89, 114} 
\xdefinecolor{tuviolet}{RGB}{61, 52, 139} 
\xdefinecolor{tumagenta}{RGB}{139, 52, 88} 
\xdefinecolor{tudarkergreen}{RGB}{75, 98, 44} 
\xdefinecolor{tuolive}{RGB}{83, 145, 45} 
\xdefinecolor{tulime}{RGB}{215, 215, 0} 
\xdefinecolor{tugraygreen}{RGB}{217,233,229} 

\newcommand{\ci}{\ensuremath{\perp\kern-5pt\perp}}

\newcommand{\bigO}{\mathcal{O}}
\DeclarePairedDelimiterX{\inner}[2]{\langle}{\rangle}{#1, #2}

\def\vec#1{\mathchoice{\mbox{\boldmath$\displaystyle#1$}}
	{\mbox{\boldmath$\textstyle#1$}}
	{\mbox{\boldmath$\scriptstyle#1$}}
	{\mbox{\boldmath$\scriptscriptstyle#1$}}}

\DeclareSIUnit{\kWh}{kWh}
\DeclareSIUnit{\MWh}{MWh}
\DeclareSIUnit{\mWs}{mWs}
\DeclareSIUnit{\Ws}{Ws}

\begin{document}
\title{The Care Label Concept:\\A Certification Suite for Trustworthy and Resource-Aware Machine Learning}
\titlerunning{A Certification Suite for Trustworthy and Resource-Aware Machine Learning}
%
\author{Katharina Morik\inst{1}
\and Helena Kotthaus\inst{1}
\and Lukas Heppe\inst{1}
\and Danny Heinrich\inst{1}
\and \\
Raphael Fischer\inst{1}
\and Andreas Pauly\inst{1}
\and Nico Piatkowski\inst{2}} 
%
\authorrunning{K. Morik et al.}
%
\institute{TU Dortmund University, Germany\\
\email{\{firstname\}.\{lastname\}@tu-dortmund.de}
\and
Fraunhofer IAIS, Germany\\
\email{nico.piatkowski@iais.fraunhofer.de}
}
\maketitle              
\begin{abstract}

Machine learning applications have become ubiquitous. 
This has led to an increased effort of making machine learning trustworthy.
Explainable and fair AI have already matured.
They address knowledgeable users and application engineers. For those who do not want to invest time into understanding the method or the learned model, we offer \emph{care labels}: easy to understand at a glance, allowing for method or model comparisons, and, at the same time, scientifically well-based. On one hand, this transforms descriptions 
as given by, e.g., Fact Sheets or Model Cards, into a form that is well-suited for end-users. On the other hand, care labels 
are the result of a certification suite that tests whether stated guarantees hold. 

In this paper, we present two experiments with our certification suite. 
One shows the care labels for configurations of Markov random fields (MRFs). 
Based on the underlying theory of MRFs, each choice leads to its specific rating of static properties like, e.g., expressivity and reliability. 
In addition, the implementation is tested and resource consumption is measured yielding dynamic properties. 

This two-level procedure is followed by another experiment certifying deep neural network (DNN) models. 
There, 
we draw the static properties from literature on a particular model and data set. At the second level, experiments are generated that deliver measurements of robustness against certain attacks. We illustrate this by ResNet-18 and MobileNetV3 applied to ImageNet.



\end{abstract}

\section{Introduction}
Since machine learning (ML) has become a prime enabling technology for diverse application areas, several academic fields and society discuss capabilities and limitations of artificial intelligence (AI). Aspects range from safety
\cite{Houben/etal/2021a}, accountability 
\cite{Bellotti/Edwards/2001a}, and responsibility
\cite{Dignum/2019a} 
to general ethical questions
\cite{Floridi/etal/2018a}.
Trustworthy AI answers the societal demands
\cite{Braunschweig/Ghallab/2021a/fixed} and surveys methods for increasing trustworthiness 
\cite{huang2020survey} or proposes AI internal audits
\cite{Raji/etal/2020a}.
Since there are diverse related parties in each ML application, diverse methods are needed for each stakeholder type
\cite{Langer/etal/2021a}.
Explanations of methods or learned models allow application specialists a detailed interaction with system developers
\cite{Guidotti/etal/2018a,Samek/etal/2019a}.
Recently, the need of documenting interactions with respect to fairness has been put forward
\cite{Sokol/Flach/2020a}.
Approaches to auditing and explanations are important for interested parties in the application area and society. Interactive modeling of applications has a long tradition within ML
\cite{Morik/94a}.
All these attempts demand a considerable amount of preoccupation with ML methods and models, however not everybody has the time or is inclined to dig that deep.
Hence, in this paper, we describe a different undertaking.

\begin{figure}[t]
    \centering
    \includegraphics[width=.95\textwidth]{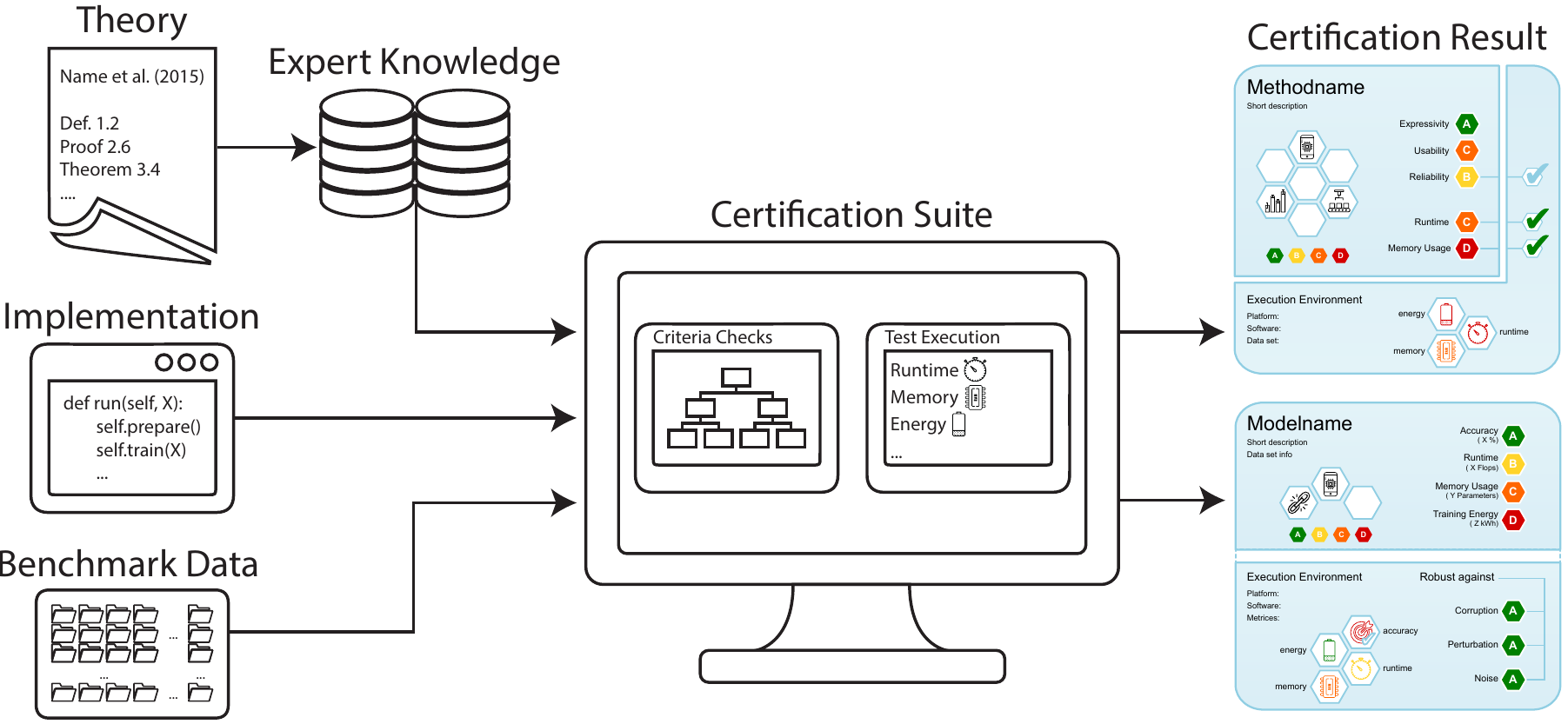}
    \caption{Certification suite for trustworthy and resource-aware machine learning}
    \label{fig:framework}
\end{figure}

For users who want to deploy a learned model in a similar way as they use their kitchenware or electrical equipment, we adapt \emph{care labels} \cite{morik2021yes,Chatila/etal/2021}, 
that turn knowledge about ML systems into guarantees for their use.
We enhance the approach via certification tests. 
Our framework is schematically displayed in \autoref{fig:framework}.
On the one hand, care labels may be considered smart transformations of ML process descriptions or documentations, that are customary.
ML pipelines together with all hyperparameter settings are visually presented at different levels of abstractions by tools like, e.g., RapidMiner
\cite{Mierswa/etal/2006a}, where also short characteristics of used methods are available as text.
For \acp{dnn}, which are pipelines in their own right, Fact Sheets
\cite{Richards/etal/2020a} and 
Model Cards
\cite{Mitchell/etal/2019a} 
allow for documentation of learned models.
Hyperparameters of training experiments and the model together with training and testing data are described in textual form.
Several repositories store experimental results and corresponding code, data and learned model\footnote{\url{https://paperswithcode.com}}. This is all made for developers and, indeed, we also use these sources of knowledge about models.
In contrast, we construct ratings of properties according to literature and communicate them in form of \textit{A}(best) to \textit{D}(worst).

On the other hand, care labels express properties based on underlying theory of ML methods. Each class of methods introduces specific categories that assign proven properties to its algorithms for training and inference.
There are several algorithms for learning decision trees, support vector machines, K-Means clustering, to name just a few. For many variants, theoretical results 
exist that prove potentials and limitations of the method. 
Representing this knowledge in the form of meta-data allows for an automated characterization and rating, which is presented by the care label.
In this paper, the class of exponential families illustrates our approach.
It shows how we exploit configuration options of a method.
A \ac{mrf} is configured by choice of a loss function, inference algorithm, and optimizer.
Each configuration yields a specific care label based on underlying theory, we show them for \acp{mrf} in \autoref{sec:experiments}.

Finally, the static characteristics of a method or model are not enough.
It is necessary to test, whether the properties hold for a given implementation.
Here, we enter the field of certification~\cite{Cremers/etal/2019a}, especially that of verifiable claims~\cite{Brundage/etal/2020a}. 
Moreover, we want to also take into account the hardware platform.
Particular properties refer to algorithms tailored to, e.g., CPU, GPU, microcontrollers, ultra-low power devices
\cite{Piatkowski/etal/2014a} or low voltage accelerators
\cite{Stutz/etal/2020a}.
In addition to runtime and memory usage, energy consumption is measured and shown in the care label.
For \acp{dnn}, robustness tests are categories of the dynamic care label part.
The contributions of this paper are:
\begin{itemize}
\item Novel presentation format for ML descriptions 
\begin{itemize}
    \item easy-to-understand care label design,
     \item criteria together with their rating in the respective value range.
\end{itemize}
\item Theory-based characteristics of ML methods
\begin{itemize}
    \item exploit theoretical guarantees of ML algorithms, if tight bounds exist,
    \item exploit published results of models on data sets, otherwise.
\end{itemize}
\item Testing properties of ML
\begin{itemize}
    \item the characteristics are written in the form of meta-data and
    \item meta-data drive the certification procedures.
\end{itemize}
\end{itemize}
In this paper, we report on related work in \autoref{sec:sota}, before we describe our approach in \autoref{sec:methods}. Extensive experiments with both \acp{mrf} and \acp{dnn} (\autoref{sec:experiments}) show the broad range already covered. Section \ref{sec:concl} concludes. 

\section{Related Work}
\label{sec:sota}




The call for explainability has come along with the success of \acp{dnn}, cf. e.g., \cite{Samek/etal/2019a}. Many approaches exist for interaction of model developers and model deployers. Care labels summarizing properties at a glance, however, need some verification of properties. In the following, we first report on work on which we base care labels for \acp{dnn}. Second, we report on work allowing care labels for \acp{mrf}, since we should not forget about probabilistic or information-theoretic methods.

\subsection{Verification and Testing of DNN}
\acp{dnn} as universal function approximators are composed of layers, activation functions, and are ultimately defined by a large number of hyperparameters. Resulting prediction quality or robustness, in the contrary, is not compositional. Theoretical bounds on the sample complexity or learnability in the PAC framework often refer to the VC-dimension. Recently, tighter bounds of the VC-dimension of \ac{dnn} have been proven for those using only the ReLU activation function
\cite{Harvey/etal/2017a}. Similarly, for ReLU networks, a verification procedure has been proposed using satisfiability modulo theories (SMT)
\cite{Katz/etal/2017a}. 
They do not take into account, whether it is a convolutional \ac{dnn}, a variational auto-encoder, or even a generative adversarial network (GAN). 

While theoretical constructs such as neural tangent kernels~\cite{Jacot/etal/2018a} or eigenvalue computations~\cite{Xiao/etal/2020a} help in understanding the theoretical properties, they cannot yet be used to assign care labels to the variants of \acp{dnn}.
The survey on verification and testing of \acp{dnn}
\cite{huang2020survey} presents SMT solvers and tests that are based on efficiently determining interesting regions 
\cite{huang2017safety}.
Reachability of true target values can be guaranteed through nested optimization\cite{Ruan/etal/2018a}.
General bounds do not differ for different types of \acp{dnn}, i.e. we cannot derive distinguishing property values from these theoretical results for classes of \acp{dnn}. More importantly the complexity of calculating properties based on this theory is prohibitive.

Hence, we do not assign care labels to particular types of the \ac{dnn} method. Instead, we have to base our care labels for \acp{dnn} on trained 
\emph{models} of particular architectures and data sets. This is very close to Fact Sheets \cite{Richards/etal/2020a} and 
Model Cards
\cite{Mitchell/etal/2019a}, but different in the presentation and verification of properties. 
Testing methods for trained models based on neuron coverage
\cite{HarelCanada/etal/2020a} or 
benchmarking
\cite{Hendrycks/Dietterich/2019a} are useful.
Research has focused on verifying robustness and generalization capabilities of \acp{dnn} using carefully crafted corruptions and attacks specifically tailored towards a single use case.
Szegedy et al. \cite{Szegedy/etal/2014a} have shown that considering robustness against small perturbations, so-called adversarial attacks, is crucial in the verification process.
A game-theoretic approach to achieve guarantees for \acp{dnn} is based on a maximum safe radius regarding the attack
\cite{Wu/etal/2020a}.
Types of adversarial attacks are introduced by
\cite{Kurakin/etal/2017a,Tramer/Boneh/2019a,Chen/etal/2020a}, 
and provided by tools like CleverHans \cite{Papernot/etal/2018a}. 

The energy consumption especially for training a \ac{dnn} is an important criterion and needs to be shown on a care label. The estimated CO2 consumption has been listed for NLP training
\cite{Strubell/etal/2020a}.
On the one hand, energy-efficient architectures are needed for model deployment on, e.g., mobile phones and trained models are specialized for different hardware architectures
\cite{Cai/etal/2020a}. 
On the other hand, methods for estimating the energy consumption for deep learning are proposed  
\cite{Martin/etal/2019a,Henderson/etal/2020a}.
For the care labels, we make good use of available tools. 

\subsection{Probabilistic Graphical Models}

In this paper, we work with \acp{mrf} which 
are theoretically well-founded probabilistic graphical models. 
Their generative nature allows utilization for complex tasks \cite{Fischer/etal/2020b}, and they have been adapted for distributed learning \cite{Heppe/etal/2020a} and for systems under tight resource constraints \cite{Piatkowski/etal/2016a}.

Given a vector of random variables $\vec{X}=(X_1, X_2, \ldots, X_n)$, \acp{mrf} model the joint probability density and can be expressed according to the Hammersley-Clifford theorem~\cite{Hammersley/Clifford/71a} by a graph $G=(V,E)$ factorizing over the set of maximal cliques:
\begin{equation}
    \mathbb{P}(X=x) = \frac{1}{Z}\prod_{C \in \mathcal{C}(G)} \psi_C(x_C) \enspace,
\end{equation}
where $\psi_c(x_c)$ is a non-negative potential function parametrized by some weights $\vec{\theta} \in \mathbb{R}^{n}$ and Z is the partition function ensuring proper normalization
\begin{equation}
    Z = \sum_{x \in \mathcal{X}} \prod_{C \in \mathcal{C}(G)} \exp{(\langle \vec{\theta}_C, \phi(x)_C\rangle}).
\end{equation}
Here we set $\psi_c(x_c) = \exp{(\langle \vec{\theta}_c, \phi(x)_c\rangle)}$ to the canonical exponential family with clique-wise one-hot encoded sufficient statistics $\phi(x)$.
Since $\mathbb{P}$ factorizes over the cliques, $\vec{\theta}$ is the concatenation of individual clique parameters.

Given a data set $\mathcal{D}$, weights $\vec{\theta}$ can be estimated by specifying a loss function and an optimizer, e.g., first-order gradient-based methods on log-likelihood. 
Therefore, an optimizer and a loss function have to be specified. 
Common choices are gradient descent and likelihood function. 
Given a set of parameters $\vec{\theta}$ we can perform probabilistic inference, e.g., perform queries for marginal probabilities or maximum-a-posteriori states.
Probabilistic inference is \#P-complete in general. 
A prominent representative of exact methods is the \ac{jt} algorithm~\cite{Lauritzen/Spiegelhalter/88a}, which yields exact results on arbitrary graphs at cost of exponential runtime and memory cost in size of the largest clique.
To overcome this limitation, several algorithms for approximate inference such as \ac{lbp} were proposed~\cite{Murphy/2013a}, which trade guarantees against saving resources.   
As we shall see, \ac{mrf} care labels benefit from theoretical results.

Early model-checking approaches for verifying the accuracy of probabilistic models have been presented 
\cite{Kwiatkowska/etal/2011}. These refer to other types of models, namely 
Markov chains,  
    Markov decision processes, and 
    probabilistic automata. 
Further work could use the insights there for another set of care labels. 


\renewcommand\tabularxcolumn[1]{m{#1}}

\section{Care Label Certification Suite}
\label{sec:methods}

For certifying the behavior of ML systems and inform end-users about their properties, we propose a certification suite, visualized in \autoref{fig:framework}.
It draws from three inputs: the expert knowledge database, a given ML implementation, and benchmark data.
The expert knowledge consists of \emph{static} criteria based on scientific work and theoretical results, both are used to generate tests and ratings (criteria checks).
Furthermore, the implementation and benchmark data are used to run performance tests and assign ratings to \emph{dynamic} properties, e.g., energy consumption per prediction (test performance).
Following this two-level procedure, the suite provides the user with care labels, which summarize the findings and test results for certification and comparison.
We start by discussing our design concept for care labels.
We then explore how their content is generated and tested, namely the static (\autoref{ssec:static_properties}) and dynamic properties (\autoref{ssec:dynamic_properties}).

\subsection{Design Concept}

We propose two care label designs as displayed in \autoref{fig:framework}, one for particular trained models (\emph{models} for short), and one for the methods with tight bounds that can be efficiently computed (\emph{methods} for short).
Our design allows non-experts to understand and compare different ML systems regarding their trustworthiness and resource consumption.
Care labels are subdivided into two segments, informing about static (\emph{Method-, Modelname}) and dynamic (\emph{Execution Environment}) properties. 
Inspired by well-known certificates such as energy or Nutri-score labels, we denote to what extend certain properties are exhibited via an easy-to-understand four level rating system \emph{A} (best) -- \emph{D} (worst).
Similar to textile care labels, our labels display intuitive badges for noteworthy properties, e.g., a badge shaped like a \emph{mobile phone} indicating applicability for resource-constrained systems.

The upper parts include meta information and ratings that can either be statically computed or extracted from literature. 
Information in the lower parts (e.g., robustness or memory consumption) stem from executing the implementation of the method or model on a specific hardware platform.
Thus, the content of the lower parts might change dependent on the execution environment. 
The certification suite performs static checks based on expert criteria and dynamic tests based on the implementation and its execution environment.
In the following sections, respective properties, tests, and ranking procedures are introduced.

\subsection{Static Properties: Expert Criteria}
\label{ssec:static_properties}
\begin{figure}
     \centering
     \includegraphics[width=\textwidth]{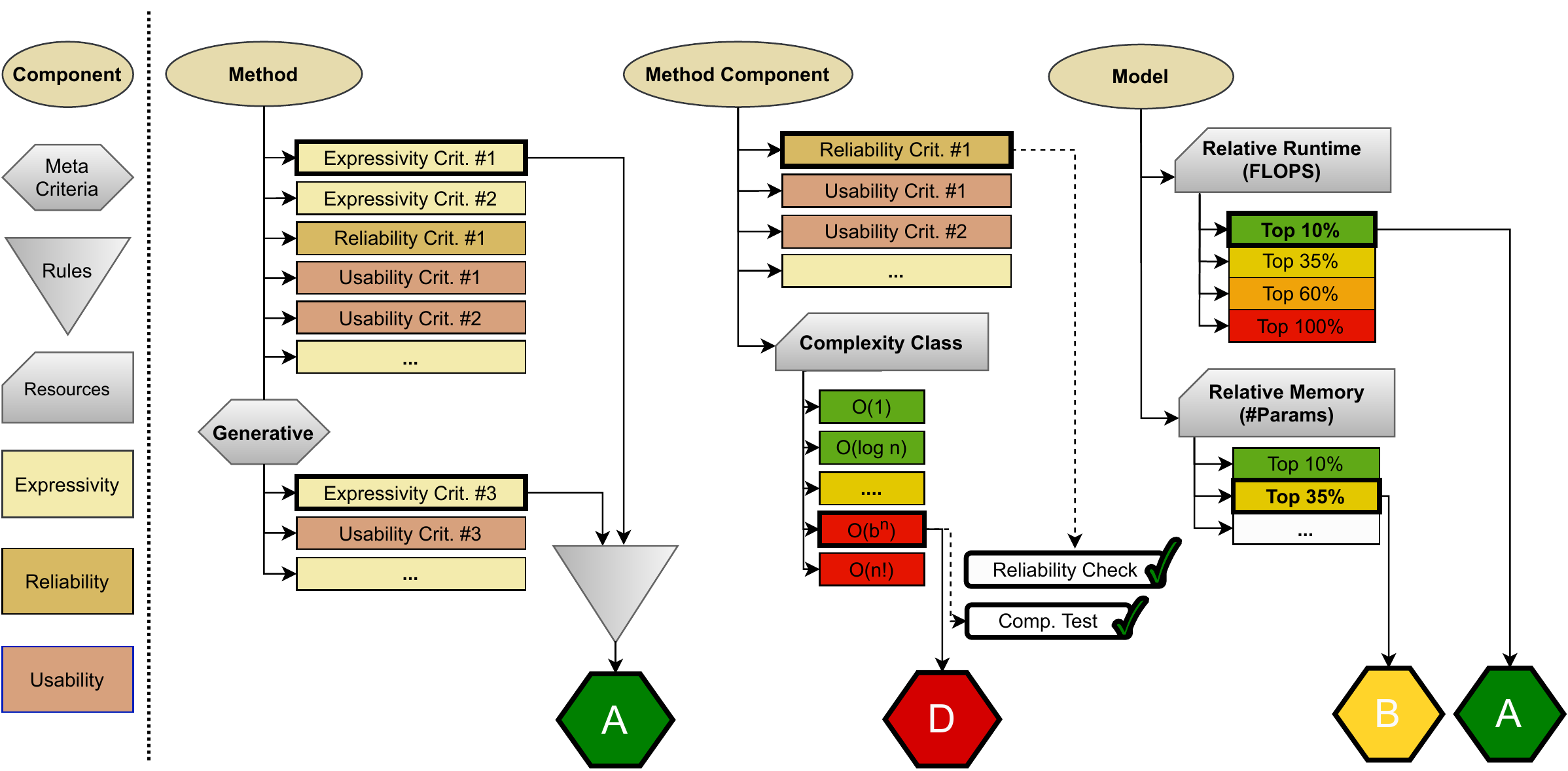}
     \caption{Schematic representation of the meta-data driven certification procedure. For different ML components, sets of expert criteria (meta-data grouped into categories) are assessed. By applying expert rules and scales to the criteria the final rating is generated.
     }
     \label{fig:meta-proc}
\end{figure}

For processing expert knowledge, we developed a meta-data driven certification procedure, visualized in \autoref{fig:meta-proc}. 
For each \emph{Component} of a method, e.g., \ac{jt} as an inference algorithm choice for discrete \acp{mrf} (see \autoref{fig:mrf_care_labels}), we denote static properties via expert criteria that
are grouped into the categories \emph{Resources}, \emph{Expressivity}, \emph{Reliability}, and \emph{Usability}.
The last three categories are rated based on composed \emph{Rules}, e.g., for simple cases as combination of fulfilled / not fulfilled criteria. 
Certain fulfilled criteria can also directly trigger \emph{Checks} or \emph{Tests} for certifying implementation properties. 

The category \emph{Resources} is either derived from theoretical asymptotic worst-case scenarios for methods or determined by a dynamic ranking.
For ML \emph{Methods} and components, the rating for runtime and memory usage, w.r.t. input data dimension is reported on a worst-case \emph{Complexity} scale (big $\mathcal{O}$-notation).
For ML \emph{Models} (see \autoref{fig:bnncnn}), the resource category is rated via a relative task-specific ranking, based on statically extracted resource requirements.
As the model complexity of \ac{dnn} architectures typically reduces to a constant, we instead report and rate the memory usage via number of \emph{Parameters} and runtime via number of \flops. 
In addition, we provide the energy consumed by a model during training to provide an idea about expenses and environmental impact.



\subsection{Dynamic Properties: Execution Environment}
\label{ssec:dynamic_properties}
The dynamic properties of ML methods and models are either certified by checks that are generated based on expert criteria, or consist of execution environment specific performance measurements.
This eases the comparison of different ML implementations regarding resource consumption.
For the ML methods (see \autoref{fig:mrf_care_labels}), green check marks denote whether theoretical bounds on resource consumption and reliability hold.
Unfortunately, checks on reliability as well as complexity class checks only partially apply to \acp{dnn}. However, the runtime and memory requirements for \acp{dnn}, especially for inference tasks, can be directly derived from the number of layers, parameters and floating point operations, independently from the execution environment. 
Instead, we verify \acp{dnn} using robustness tests previously mentioned in \autoref{sec:sota}.
For the models (see \autoref{fig:bnncnn}) the A-D ratings denote the relative performance of robustness tests.

We argue that for certification, most expert criteria that trigger implementation tests should be model or method-specific.
Thus, we present the \ac{mrf} and \ac{dnn}-specific tests in \autoref{sec:experiments}.
In the following we discuss the details of rating and testing runtime and memory complexity for ML methods. 

Runtime and memory complexity can commonly be derived from the underlying algorithmic requirements of a certain method.
While the computation complexity is usually given in worst-case $\bigO$-notation, implementations may incur large coefficients on the asymptotic runtime.
However, it is also possible that implementations may not adhere to the given asymptotic runtime due to overhead or wrong implementation of an algorithm.
To \emph{verify runtime and memory costs} one can scale the given $n$ to obtain a set of experiments that can be used to fit a set of curves e.g. linear $f(x) = ax + b$, quadratic $f(x) = ax^2 + bx + c$ and exponential functions $f(x) = a \cdot \exp{(b x)} + c$ to obtain the curve that contributes most to fitting the original memory or runtime data $y$ for some scaling input $x$.
Here we minimize the error for a set of selected complexity classes $\mathcal{F}$:
\begin{equation}
    \arg\min_{f\in \mathcal{F}} \bigg\{\min_{\vec{\theta}}   \sum_{\vec{x}\in \mathcal{D}} || y - f(\vec{x}, \vec{\theta})||_2^2+ \lambda R(\vec{\theta})\bigg\}
\end{equation}

This check is used in \autoref{ssec:experiments_mrf} to verify the implementation's theoretical runtime and memory consumption.

\section{Experiments: Applying the Care Label Certification} \label{sec:experiments}




In this section, we apply our certification suite to \acp{dnn} and \acp{mrf} in order to illustrate our approach. We combine the robustness tests for DNNs mentioned in \autoref{sec:sota} with our general concept presented in \autoref{sec:methods} to showcase the care label concept.


\subsection{Deep Neural Networks}
\label{sec:DNN}
In our experiments we focus on inference tasks, as the application of learned models is a common entry point for developing or extending new use cases. 
Properties of a neural network depend on the specific model architecture and thus also on the data set. 
Therefore, \ac{dnn} care labels should always be issued for a specific combination of data set and model architecture. 

For our experiments, we chose the PyTorch framework, since it provides pre-trained models via ModelZoo\footnote{\url{https://pytorch.org/serve/model_zoo.html}} together with the popular \texttt{ImageNet} data. 
To allow comparison and relative ranking, we ran experiments with four renowned models, namely \alexnet~\cite{Krizhevsky/etal/2012a}, \vgg~ \cite{Simonyan/Zisserman/2015a}, \resneteighteen~\cite{He/etal/2016}, and \mobilenetlarge~\cite{Howard/etal/2019a}.
For the two most recent models (\resneteighteen and \mobilenetlarge), the final care labels are depicted in \autoref{fig:bnncnn}.
Note that more extensive experiments on a wider range of models would improve and refine our care label ranking results.

\subsubsection{Static Properties}
In the upper segments of the care label (see \autoref{fig:bnncnn}), statically extracted information like accuracy, number of giga\flops (runtime), number of parameters (memory), and training energy consumption is presented for specific model architectures. To calculate the number of {G\flops} and parameters we used the \texttt{thop}\footnote{\url{https://github.com/Lyken17/pytorch-OpCounter}} software package. 
Since the training energy cannot always be directly extracted from literature, e.g., via number of training epochs, we measured average wattage and epoch time for the respective models over ten epochs. 
These values are then multiplied with the number of epochs, required to reach target accuracy according to \texttt{TorchVision}\footnote{\url{https://github.com/pytorch/vision/tree/master/references/classification}}.


\begin{table}[h]
    \centering
    \resizebox{\textwidth}{!}{
        \begin{tabular}{l|c|c|c|c|c}  
        \toprule 
        Model (ImageNet) & Acc@Top1 & Acc@Top5 & Runtime (GFLOPS) & Memory (M Parameters) & Train Energy (\si{\kWh})  \\
        \midrule
        \midrule
        AlexNet \cite{Krizhevsky/etal/2012a} & 56.52 & 79.07 & 1.43 & 61.10 & \textbf{2.47} \\
        VGG11 \cite{Simonyan/Zisserman/2015a} & 69.02 & 88.63 & 15.23 & 132.86 & 18.12 \\
        ResNet-18 \cite{He/etal/2016} & 69.76 & 89.08 & 3.64 & 11.69 & 6.55 \\
        \mobilenetlarge \cite{Howard/etal/2019a} & \textbf{74.04} & \textbf{91.34} & \textbf{0.45} & \textbf{5.48} & 35.13 \\
        \bottomrule
    \end{tabular}}
    \caption{Static properties for different DNN architectures using ImageNet. For each column the best value is highlighted (rating \emph{A}).}
    \label{tab:dnn-static}
\end{table}


For the labels, we derive rankings (\emph{A}-\emph{D}) via relative comparison of static model information.
We divide information for each property (i.e. columns of \autoref{tab:dnn-static}) into four quantiles and rate a given model based on the quantile it falls into.
Thus, we award \mobilenetlarge with an \emph{A} for runtime, memory, and accuracy, scoring best in model comparison (\resneteighteen only receives \emph{B} and \emph{C} ratings).
The labels however also indicate the downside of \mobilenetlarge, as training energy is rated with a \emph{D}.
In fact, the required amount of energy is more than five times higher compared to the 6.55 \si{\kWh} of \resneteighteen (awarded a \emph{B}), due to the higher number of epochs (90 vs 600 epochs).
The big advantage of \mobilenetlarge is its suitability for mobile processors due to its modest parameter footprint, which the label illustrates via the \emph{mobile phone} badge.

\subsubsection{Dynamic Properties}
\begin{figure}[t]
    \centering
    \includegraphics[width=.98\textwidth]{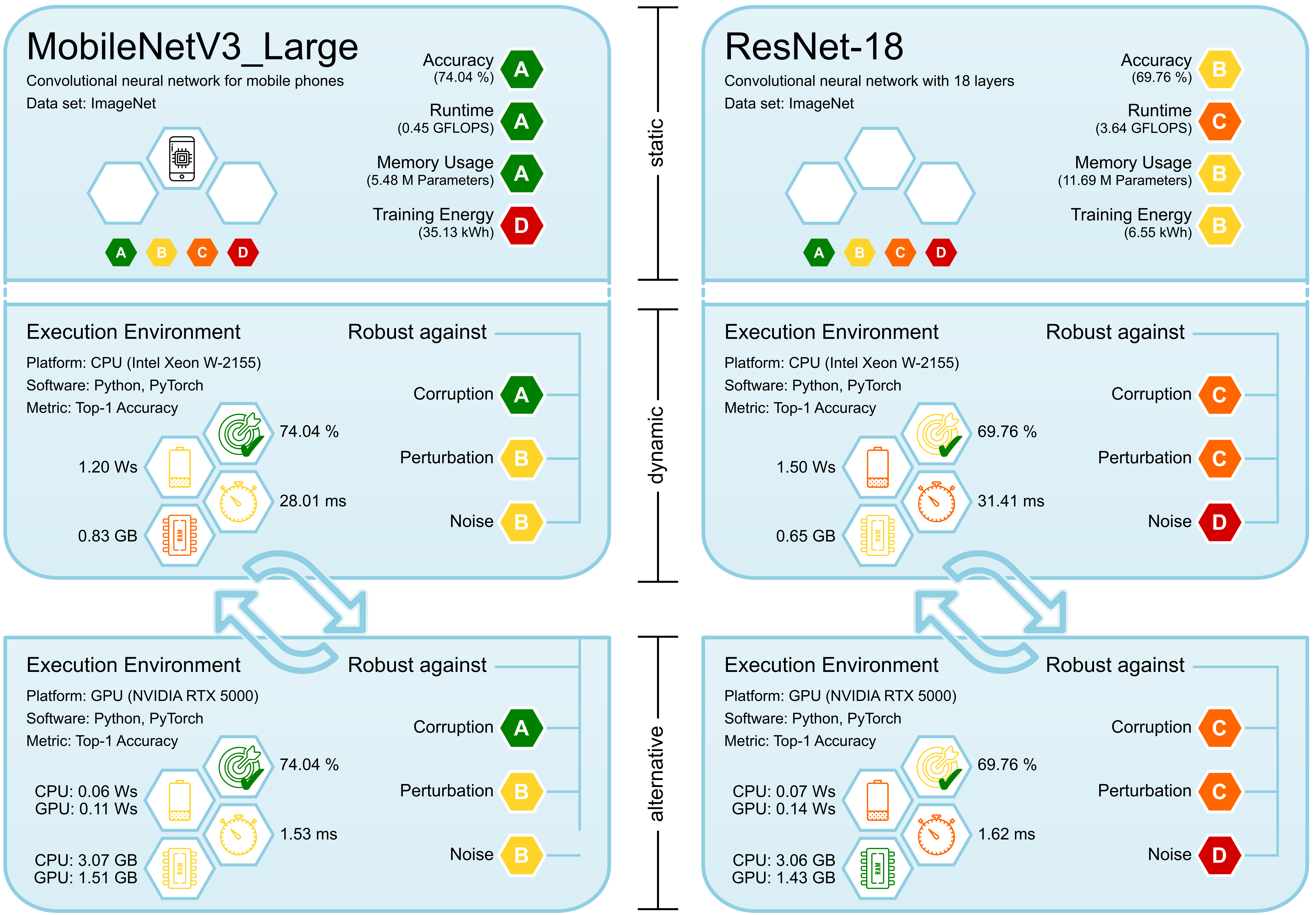}
    \caption{Care labels for \mobilenetlarge and \resneteighteen  on different platforms.}
    \label{fig:bnncnn}
\end{figure}

We investigate dynamic properties of the previously mentioned models during inference on different execution environments for CPU and GPU.
All experiments have been executed on the same system, running Ubuntu 20.04 as operating system, equipped with an Intel Xeon W-2155 \cpu,  64 GB of main memory, and an NVIDIA RTX 5000 \gpu.
For results on different execution platforms we use either the CPU alone or in combination with GPU acceleration.
\autoref{tab:dnn_results} and \autoref{fig:bnncnn} depict our findings, with runtime and energy consumption per predicted sample as well as maximum memory usage for a 3000 sample subset of ImageNet.

\begin{table}[t]
    \centering
    \resizebox{\textwidth}{!}{
    \begin{tabular}{l|c|c|c|c|c|c|c|c|c}  
    	\toprule 
    	\multirow{2}{*}{Model (ImageNet)}  & \multicolumn{3}{c|}{Robustness} & \multicolumn{2}{c|}{Runtime ($\si{\milli\second}$)} & \multicolumn{2}{c|}{Memory ($\si{\giga\byte}$)} & \multicolumn{2}{c|}{Energy ($\si{\Ws}$)}  \\  \cline{2-10}
    	& Corr & Pert & Noise & CPU & CPU+GPU & CPU & CPU+GPU & CPU & CPU+GPU \\ 
    	\midrule
    	\midrule
        AlexNet \cite{Krizhevsky/etal/2012a} & 1.00 & 0.11 & $0.24e^{-6}$ & \textbf{10.52} & \textbf{1.24} & \textbf{0.65} & 3.20+1.57 & \textbf{0.51} & \textbf{0.05+0.07} \\
        VGG11 \cite{Simonyan/Zisserman/2015a} & 1.18 & \textbf{0.06} & $\mathbf{15.62e^{-6}}$ & 122.95 & 2.36 & 1.73 & 3.49+3.11 & 5.99 & 0.10+0.32 \\
        ResNet-18 \cite{He/etal/2016} & 1.02 & 0.08 & $0.12e^{-6}$ & 31.41 & 1.62 & 0.65 & \textbf{3.06+1.43} & 1.50 & 0.07+0.14 \\
        \mobilenetlarge \cite{Howard/etal/2019a} & \textbf{0.94} & 0.07 & $0.49e^{-6}$ & 28.01 & 1.53 & 0.83 & 3.07+1.51 & 1.20 & 0.06+0.11 \\
        \bottomrule
    \end{tabular}}
    \caption{Dynamic properties for various neural networks using ImageNet. For each column the best value is highlighted (rating \emph{A}).}%
    \label{tab:dnn_results}
\end{table}


In addition, we performed a series of \emph{robustness tests} using a set of task-specific attacks, as well as generic adversarial attacks.
As task-specific attacks, we chose the ImageNet-C and ImageNet-P benchmarks, proposed by Hendrycks and Dietterich \cite{Hendrycks/Dietterich/2019a}, where different \emph{Perturbations} and \emph{Corruptions} were applied at various severity levels. 
Additionally, we launched generic \ac{pgd} attacks, declared as \emph{Noise} on our care labels.
As proposed in~\cite{Hendrycks/Dietterich/2019a} we evaluate our models using the relative mean corruption error (rel. mCE) on ImageNet-C and flip rate (FP) on ImageNet-P.
Briefly, the rel. mCE measures the performance degradation when encountering different types of corruptions $c$ at various severity levels $s_c$. It is computed as the average performance of a model $f$ compared to its clean error $E^f_{clean}$ normalized by some baseline $b$ to allow for a fair comparison of different models, formally:

\begin{equation}
\label{eq:rel_mce_error}
    \text{rel. mCE} = \frac{1}{k}\sum_{c=1}^k \frac{1}{s_c} \frac{\sum_{i=1}^{s_c} E_{i,c}^f - E_{clean}^f }{\sum_{i=1}^{s_c} E_{i,c}^b - E_{clean}^b}.
\end{equation}



The ImageNet-P benchmark is evaluated on a series of frame sequences. Later frames of a sequence were subjected to stronger perturbations. This benchmark is evaluated via the flip rate (FP), which is defined as the empirical probability of a frame to be classified differently than its preceding frame, i.e., labels flipping between two frames:
\label{eq:flip}
\begin{equation}
    FP = \frac{1}{m(n-1)} \sum_{i=1}^m\sum_{j=2}^n \mathbbm{1}_{x^i_{j-1} \neq x^i_{j}}\\
\end{equation}


To verify robustness against noise, we use the proposed multi-step procedure proposed by~\cite{Madry/etal/2019a} and provided by~\texttt{CleverHans}~\cite{Papernot/etal/2018a} to create adversarial examples.
\begin{equation}
\label{eq:advers} 
x^{t+1} = \Pi_{x+\epsilon} (x^t + \epsilon \cdot sgn(\nabla_x L(\theta, x, y)))
\end{equation}
We estimate $\arg\max_{\varepsilon}$ w.r.t to the $L_{\infty}$ norm, such that $f(x + \epsilon) = f (x)$ for most $x \in D$.
The estimation is performed similar to binary search, for a given $\epsilon$ we generate a set $S$ of adversarial examples according to \autoref{eq:advers} and evaluate the accuracy drop $acc_{drop}$ from $f$ on $S$.
If the accuracy falls below a certain threshold, the value of $\epsilon$ is halved until convergence.
In our experiments we used $10$ steps, a threshold of $10^{-4}$, and $\epsilon = 0.001$ as initial value for generating adversarial examples (\autoref{eq:advers}).
We present all dynamically measured results in \autoref{tab:dnn_results}, and determine relative care labels ratings in the same way as for the static properties.
The robustness results for \resneteighteen indicate that it should not be used in safety-critical areas (ratings \emph{C}-\emph{D}), because predictions are unstable against corruptions, perturbations, and noise. 
In contrast, \mobilenetlarge is less vulnerable (ratings \emph{A}-\emph{B}), but also consumes more energy and memory. 
This is counterintuitive with respect to its static properties, which probably stems from using the pure PyTorch model, without any platform-related enhancements.
We thus see that static properties like G\flops and number of parameters do not act as a clear indicator on how implementations behave dynamically \cite{DBLP:conf/cvpr/TanCPVSHL19}.

Based on our ratings for robustness, the end-user gets important insights that \acp{dnn} suffer from high vulnerability against environmental influences and therefore shouldn't be applied in safety-critical areas.
%
Thus, care labels allow end-users to quickly analyze possible opportunities and challenges when applying trained models.




\subsection{Markov Random Fields}\label{ssec:experiments_mrf}
We provide care labels for \acp{mrf} in combination with either \ac{jt} inference or \ac{lbp} to highlight the implications caused by the algorithm choice. 
Our results in the form of care labels are shown in \autoref{fig:mrf_care_labels}. 

\subsubsection{Static Properties}

MRFs receive an \emph{A} for \emph{Expressivity} in both care labels (see \autoref{fig:mrf_care_labels}, upper part), since they can model any distribution of the exponential family. 
Besides, they provide uncertainty estimates and allow to solve a variety of probabilistic inference tasks. 
For \emph{Usability}, the JT care label is awarded with \emph{B}, while LBP is rated \emph{C}.
The reasoning here is that the underlying graph structure has to be either known or estimated for applying MRFs.
The LBP label was assigned a lower rating due to requiring additional hyperparameters, namely convergence criteria for performing probabilistic inference. 
In terms of \emph{Reliability}, an \emph{A} was determined for the JT configuration, since it is an deterministic algorithm and answers inference tasks exactly. 
As no such guarantees exist for LBP, a \emph{D} was assigned.
We here linked an underlying criteria to a reliability check, namely the \emph{Probability Recovery Check}.

Looking at resource requirements, the JT care label was assigned a \emph{D} rating for \emph{Runtime} as well as \emph{Memory Usage}, due to exponential scaling in the tree width \cite{Piatkowski/2018a}.
In contrast, the LBP algorithm is quite resource friendly, since it grows at most quadratically in the number of states or edges.
This behavior is also tested in the dynamic segment, which we now discuss in detail.



\subsubsection{Dynamic Properties}
\begin{figure}[t]
    \centering
    \includegraphics[width=\linewidth]{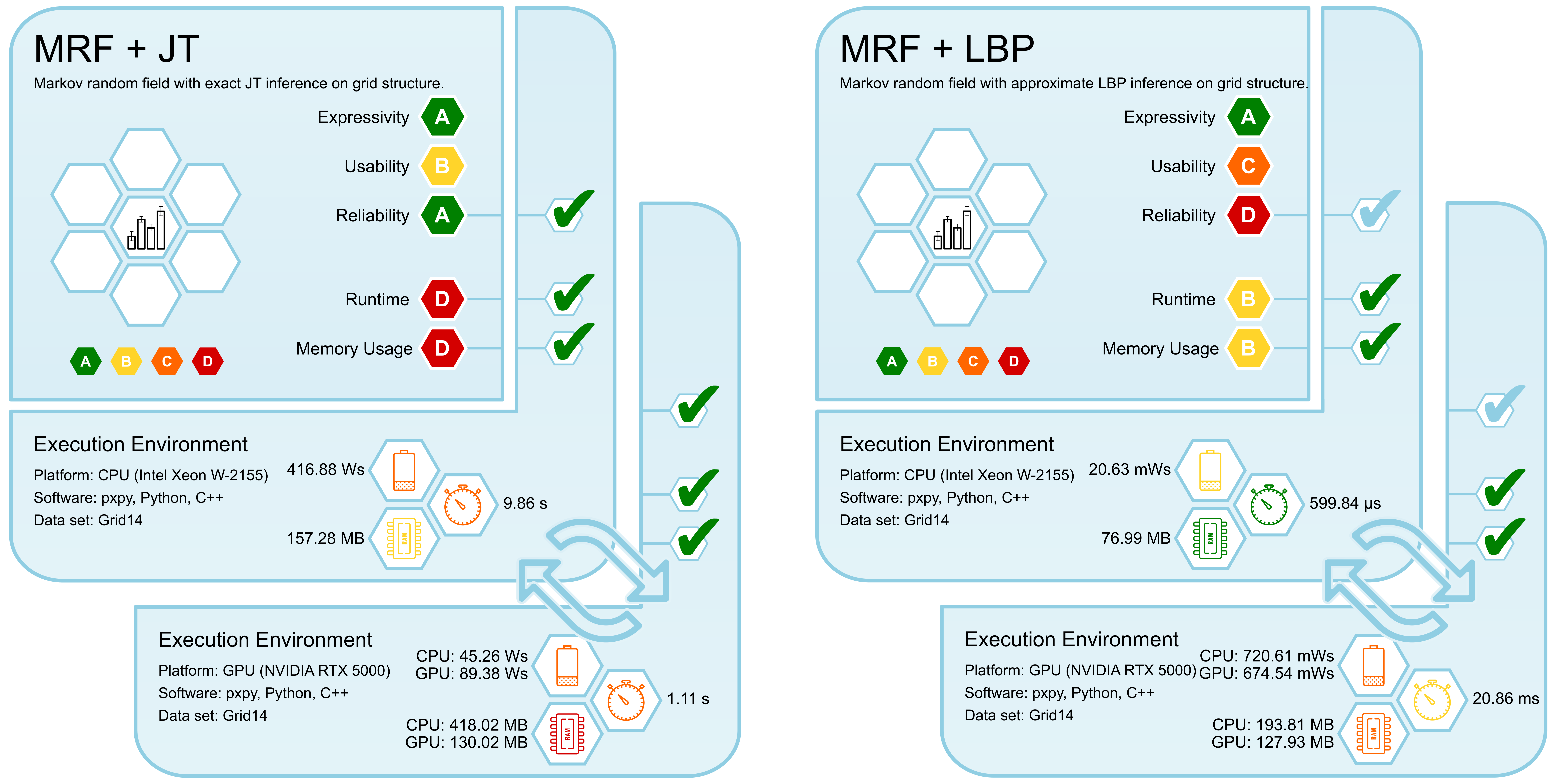}
    \caption{Care labels for \acp{mrf} with different inference algorithms and platforms.}
    \label{fig:mrf_care_labels}
\end{figure}

The dynamic results for our MRF variants are shown in the lower parts of~\autoref{fig:mrf_care_labels}, each executed on two different environments (cf. \autoref{sec:DNN}). 
As implementation we choose the \pxpy library\footnote{\url{https://pypi.org/project/pxpy/}}, since it provides implementations for different MRF configurations. 
For experiments, synthetic data was generated with grid graphs as underlying graph structure and binary states per vertex.
We performed two types of checks, complexity class checks for memory and runtime bounds and reliability checks, both rewarded with check marks.
In addition, we also report implementation depended resource requirements.

\emph{Reliability Checks}:  
For certifying whether the implementation fulfills given reliability guarantees we used synthetic data, allowing for comparing the model output against known parameters.
In the context of \acp{mrf}, we propose to perform a \emph{Probability Recovery Check}.
This test computes the true marginal distribution for given parameters by utilizing the exact JT algorithm, and then queries the provided implementation for its marginals under the given weights.
Afterwards, nodewise Kullback-Leibler divergences between the two marginal vectors are computed and reduced to their max value. If this falls below an acceptance threshold ($10e-3$), the check is considered passed.

\newcommand{\mpagesize}{0.45}
\begin{figure}[t]
\centering
    \begin{minipage}[t]{\mpagesize\linewidth}
        \centering
        \includegraphics[width=\linewidth]{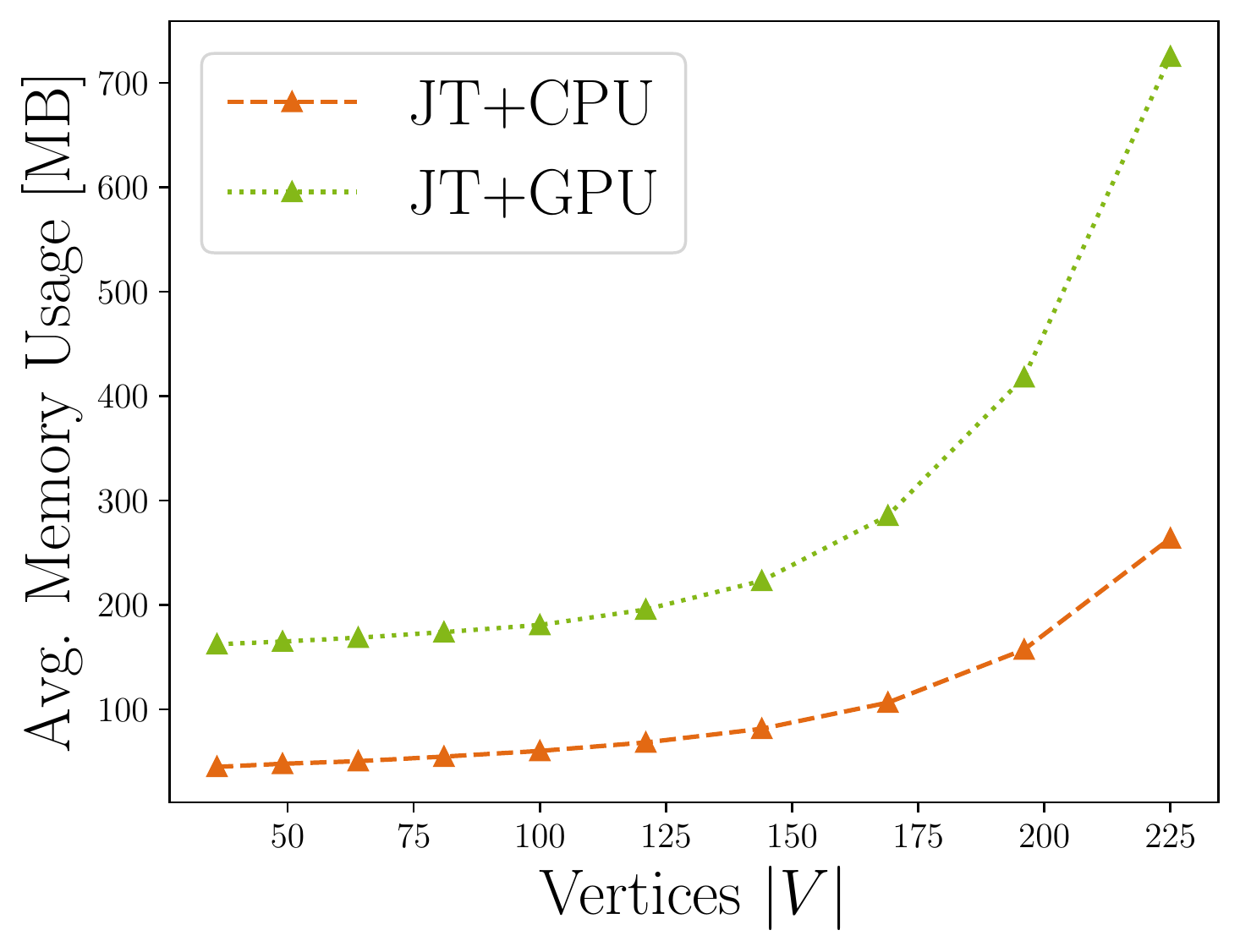}
    \end{minipage}
    \begin{minipage}[t]{\mpagesize\linewidth}
        \centering
        \includegraphics[width=\linewidth]{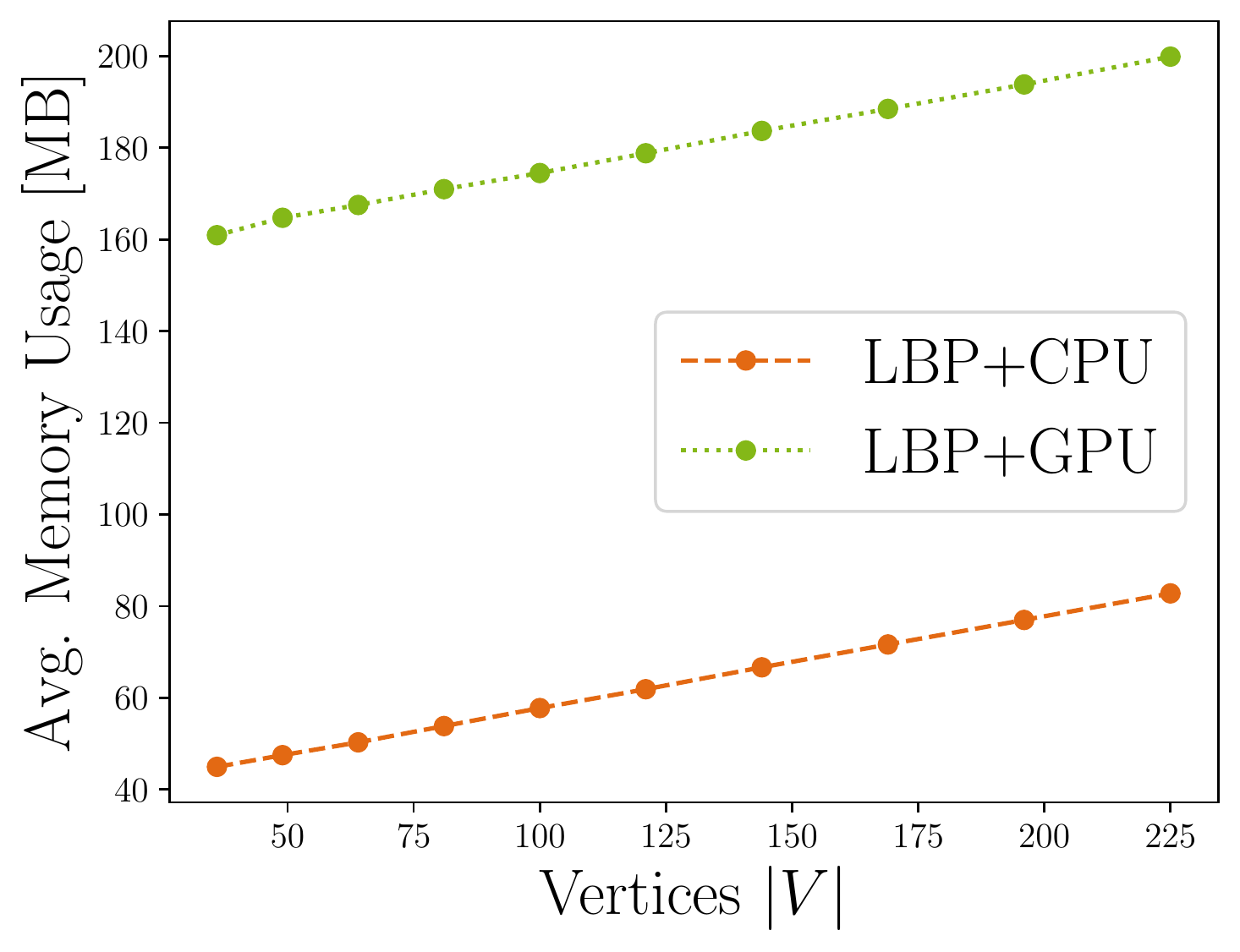}
    \end{minipage}
        \begin{minipage}[t]{\mpagesize\linewidth}
        \centering
        \includegraphics[width=\linewidth]{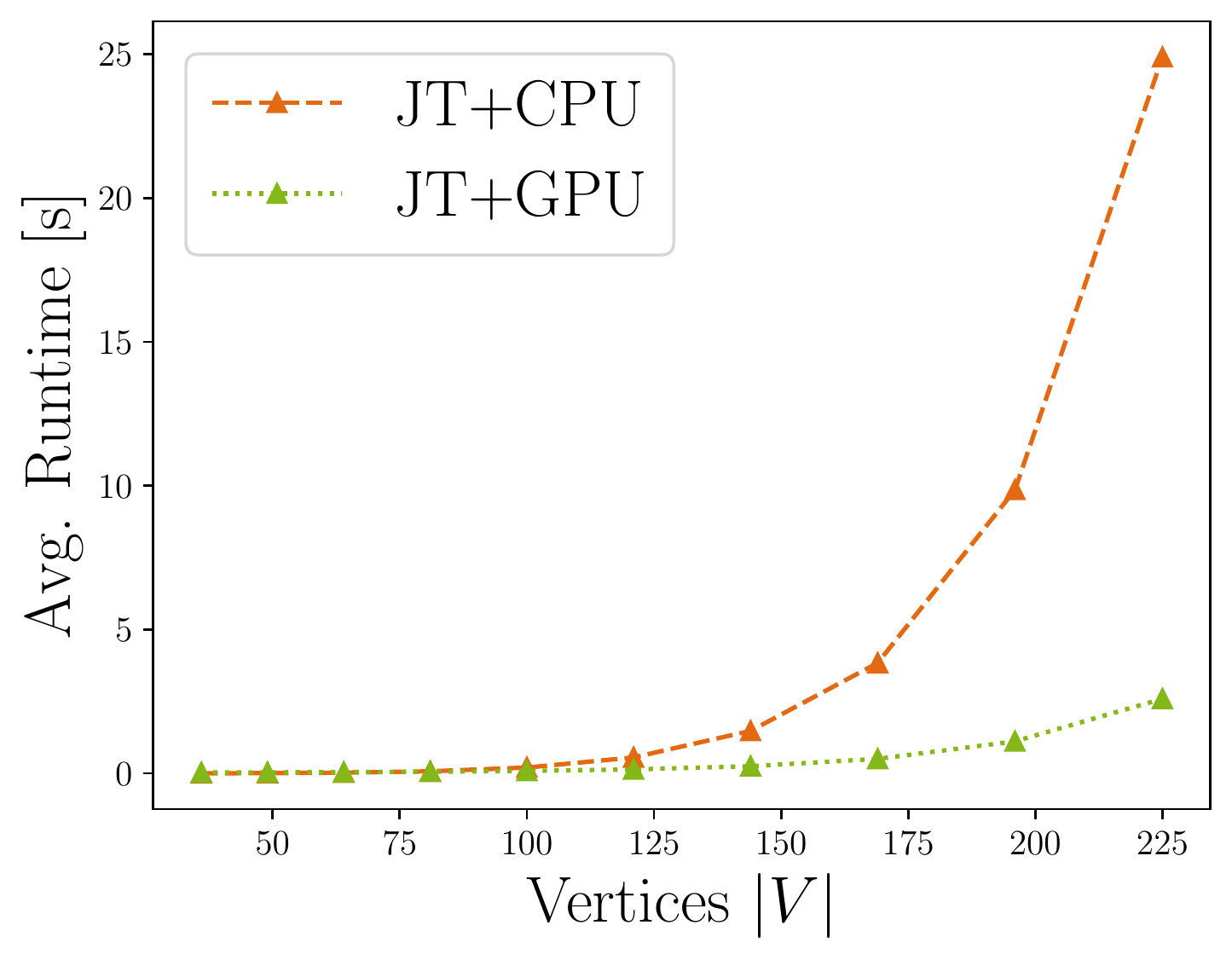}
    \end{minipage}
    \begin{minipage}[t]{\mpagesize\linewidth}
        \centering
        \includegraphics[width=\linewidth]{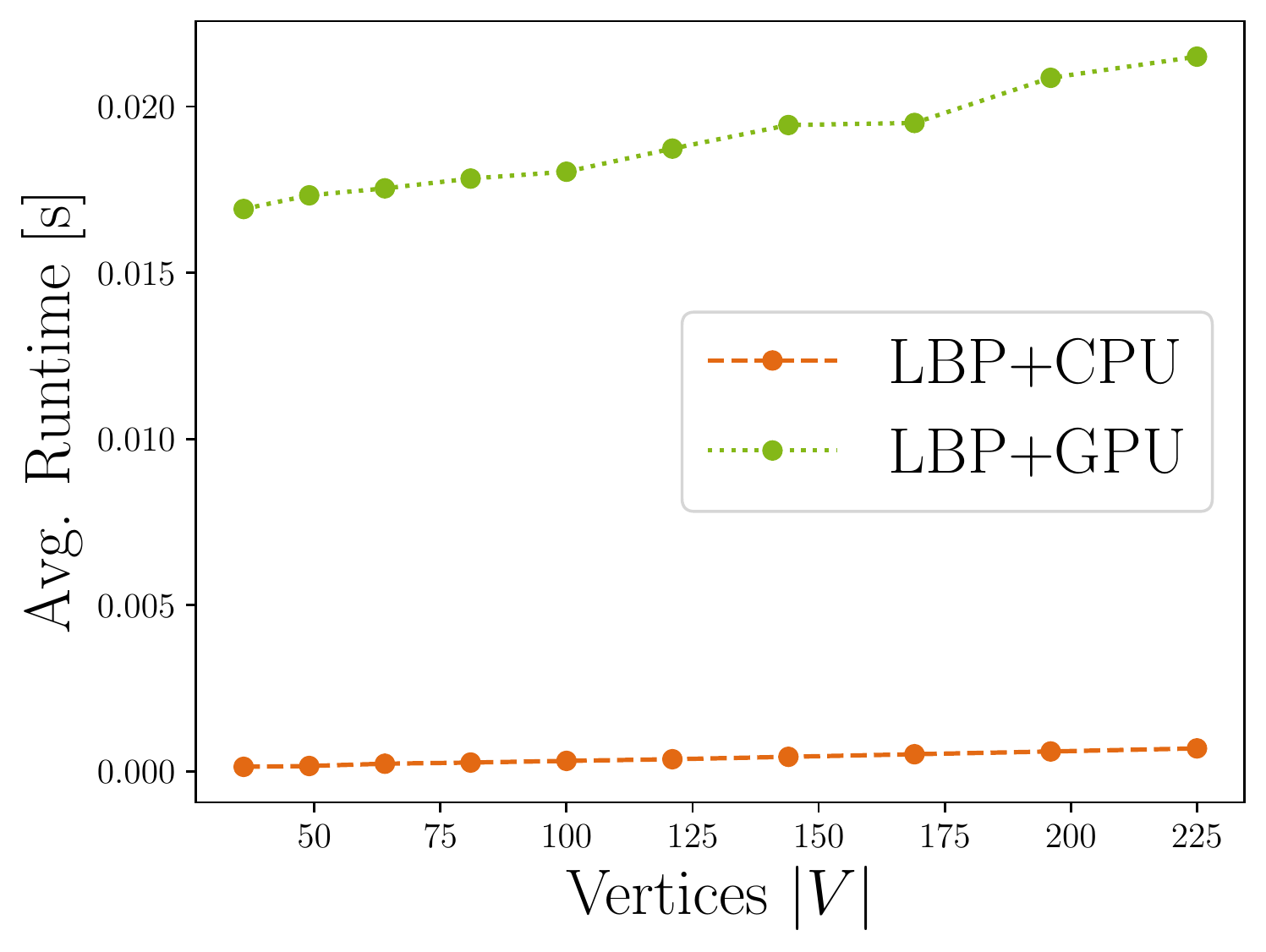}
    \end{minipage}
    \caption{We report memory [\si{\giga\byte}] and runtime [\si{\second}] requirements for a \ac{mrf} trained with \texttt{pxpy} on a grid graph with increasing grid size on GPU and CPU.}
    \label{fig:mrf_results}
\end{figure}

\emph{Complexity Class Checks} 
Here we test whether the execution behaviour on differently sized data sets scales with the theoretically given complexity classes (as explained in \autoref{ssec:dynamic_properties}).
A grid graph with nodes varying from $2 \times 2$ up to $15 \times 15$ was used as input data.
We show the underlying measurements for the runtime and memory bound checks on grid graphs in \autoref{fig:mrf_results}. 
All complexity class checks have passed, since the behavior of the algorithms was in compliance with their theoretical memory and runtime bounds.

Similar to the DNN labels, we also report general resource measurements based on a specific data set (Grid $14 \times 14$).
Here, the results highlight the differences in resource consumption with JT compared to LBP.
For the \ac{jt} algorithm we observe a trade-off between runtime and memory consumption when switching from CPU to GPU platform.
Note that for \ac{lbp} inference on GPUs, we observe a higher memory usage and runtime than for CPUs, due to I/O overhead caused by memory copy operations for small input sizes (for a larger input size this disadvantage would vanish).

\section{Conclusion}\label{sec:concl}

Today's ML technology affects diverse parties, such as application engineers and end-users.
Having only little time to thoroughly study the theory, they instead require comprehensible certificates that build trust.
Thus, a novel means of communicating information about ML is required, that has to address several points:
Firstly, it has to be easily understandable for all stakeholders, and go beyond mere model documentation.
Information on resource consumption is crucial for many application domains and needs to be transparently discussed.
Aspects of certification should ideally be based on well-established theory, such as bounds and guarantees.
Lastly, given ML implementations have to be tested, to assess their properties and check their compliance with underlying theory.

In this paper, we addressed all these aspects via care labels, obtained from our proposed certification suite.
The labels were designed to be easily understandable, hiding the underlying intricacy, and allowing for quick comparison.
We broke down the complexity of expert knowledge into meta-data criteria, enabling automated certification via label generation.
We successfully applied our concept to two popular ML representatives, namely DNNs and MRFs.
The labels for \resneteighteen and \mobilenetlarge on ImageNet show how the choice of model architecture dramatically changes properties like accuracy, robustness, or resource consumption.
For MRFs, our labels highlight the impact of choosing among probabilistic inference algorithms, which can either make the method more reliable, or more resource efficient.

This work proposed a novel format for ML certification to close the gap between experts and end-users.
There is still room for further progress, by expanding on the expert knowledge, refining how meta information is handled and processed, and of course applying the concept to more methods and models.
For making certification via care labels universally accepted, we anticipate valuable feedback from our research colleagues, and hope that they join us in the endeavour of making ML more trustworthy and resource-aware.

%
%

\subsubsection{Acknowledgement}
This research has been funded by the Federal Ministry of Education and Research of Germany as part of the competence center for machine learning ML2R (01IS18038A/B). Part of the work on this paper has been supported by Deutsche Forschungsgemeinschaft (DFG) - project number 124020371 - within the Collaborative Research Center SFB 876 "Providing Information by Resource-Constrained Analysis", DFG project number 124020371, SFB project C3.
\bibliographystyle{splncs04}
\bibliography{literature}

\begin{thebibliography}{10}
\providecommand{\url}[1]{\texttt{#1}}
\providecommand{\urlprefix}{URL }
\providecommand{\doi}[1]{https://doi.org/#1}

\bibitem{Bellotti/Edwards/2001a}
Bellotti, V., Edwards, K.: Intelligibility and accountability: Human
  considerations in context-aware systems. Human-Computer Interaction
  \textbf{16}(2-4),  193--212 (2001)

\bibitem{Braunschweig/Ghallab/2021a/fixed}
Braunschweig, B., Ghallab, M.: Reflections on Artificial Intelligence for
  Humanity. Lecture Notes in Artificial Intelligence, Springer International
  Publishing (2021)

\bibitem{Brundage/etal/2020a}
Brundage, M., et~al.: Toward trustworthy ai development: Mechanisms for
  supporting verifiable claims (2020)

\bibitem{Cai/etal/2020a}
Cai, H., Gan, C., Wang, T., Zhekai, Z., Han, S.: Once for all: train one
  network and specialise it for efficient deployment. In: Procs. {ICLR} 2020
  (2020)

\bibitem{Chatila/etal/2021}
Chatila, R., et~al.: Trustworthy AI, chap.~2, pp. 18--45. Springer
  International Publishing (2021),
  \url{https://www.springer.com/de/book/9783030691271}

\bibitem{Chen/etal/2020a}
Chen, J., Jordan, M.I., Wainwright, M.J.: Hopskipjumpattack: {A}
  query-efficient decision-based attack. In: IEEE Symposium on Security and
  Privacy. {IEEE} (2020)

\bibitem{Cremers/etal/2019a}
Cremers, A., et~al.: Trustworthy use of artificial intelligence -- priorities
  from a philosophical, ethical, legal, and technological viewpoint as a basis
  for certification of artificial intelligence. Fraunhofer Institute for
  Intelligent Analysis and Information Systems ({IAIS}) (2019),
  \url{https://www.iais.fraunhofer.de/content/dam/iais/KINRW/Whitepaper\_Thrustworthy\_AI.pdf}

\bibitem{Dignum/2019a}
Dignum, V.: Responsible Artificial Intelligence: How to Develop and Use {AI} in
  a Responsible Way. Springer (2019)

\bibitem{Fischer/etal/2020b}
Fischer, R., Piatkowski, N., Pelletier, C., Webb, G., Petitjean, F., Morik, K.:
  No cloud on the horizon: Probabilistic gap filling in satellite image series.
  In: {DSAA} 2020. pp. 546--555

\bibitem{Floridi/etal/2018a}
Floridi, L., et~al.: {AI4}people?an ethical framework for a good ai society:
  Opportunities, risks, principles, and recommendations. Minds and Machines
  \textbf{28}(4) (2018)

\bibitem{Martin/etal/2019a}
Garc{\'\i}a-Mart{\'\i}n, E., Rodrigues, C.F., Riley, G., Grahn, H.: Estimation
  of energy consumption in machine learning. Journal of Parallel and
  Distributed Computing  \textbf{134},  75 -- 88 (2019)

\bibitem{Guidotti/etal/2018a}
Guidotti, R., Monreale, A., Ruggieri, S., Turini, F., Giannotti, F., Pedreschi,
  D.: A survey of methods for explaining black box models. ACM Comput. Surv.
  \textbf{51}(5) (August 2018)

\bibitem{Hammersley/Clifford/71a}
Hammersley, J.M., Clifford, P.: Markov fields on finite graphs and lattices
  (1971)

\bibitem{HarelCanada/etal/2020a}
Harel-Canada, F., Wang, L., Ali~Gulzar, M., Gu, Q., Kim, M.: Is neuron coverage
  a meaningful measure for testing deepneural networks? In: ESEC/FSE 2020:
  Proceedings of the 28th ACM Joint Meeting on European Software Engineering
  Conference and Symposium on the Foundations of Software Engineering (November
  2020)

\bibitem{Harvey/etal/2017a}
Harvey, N., Liaw, C., Mehrabian, A.: Nearly-tight {VC}-dimension bounds for
  piecewise linear neural networks. In: Kale, S., Shamir, O. (eds.) Proceedings
  of the 2017 Conference on Learning Theory. Proceedings of Machine Learning
  Research, vol.~65, pp. 1064--1068. PMLR (2017)

\bibitem{He/etal/2016}
He, K., Zhang, X., Ren, S., Sun, J.: Deep residual learning for image
  recognition. In: {CVPR} 2016. pp. 770--778. {IEEE} Computer Society (2016)

\bibitem{Henderson/etal/2020a}
Henderson, P., Hu, J., Romoff, J., Brunskill, E., Jurafsky, D., Pineau, J.:
  Towards the systematic reporting of the energy and carbon footprints of
  machine learning. Journal of Machine Learning Research  \textbf{21}(248),
  1--43 (2020)

\bibitem{Hendrycks/Dietterich/2019a}
Hendrycks, D., Dietterich, T.: Benchmarking neural network robustness to common
  corruptions and perturbations. ICLR  (2019)

\bibitem{Heppe/etal/2020a}
Heppe, L., Kamp, M., Adilova, L., Piatkowski, N., Heinrich, D., Morik, K.:
  Resource-constrained on-device learning by dynamic averaging. In: ECML PKDD
  2020 Workshops. pp. 129--144. Springer International Publishing, Cham (2020)

\bibitem{Houben/etal/2021a}
Houben, S., et~al.: Inspect, understand, overcome: A survey of practical
  methods for ai safety. Technical Report (2021),
  \url{https://www.ki-absicherung-projekt.de/fileadmin/KI\_Absicherung/Downloads/KI-A\_20201221\_Houben\_et\_al\_-\_Inspect\_\_Understand\_\_Overcome.pdf}

\bibitem{Howard/etal/2019a}
Howard, A., et~al.: Searching for mobilenetv3. In: {ICCV} 2019. pp. 1314--1324.
  {IEEE} (2019)

\bibitem{huang2017safety}
Huang, X., Kwiatkowska, M., Wang, S., Wu, M.: Safety verification of deep
  neural networks. In: International conference on computer aided verification.
  pp. 3--29. Springer (2017)

\bibitem{huang2020survey}
Huang, X., et~al.: A survey of safety and trustworthiness of deep neural
  networks: Verification, testing, adversarial attack and defence, and
  interpretability. Computer Science Review  \textbf{37},  100270 (2020)

\bibitem{Jacot/etal/2018a}
Jacot, A., Gabriel, F., Hongler, C.: Neural tangent kernel: Convergence and
  generalization in neural networks. In: Advances in Neural Information
  Processing Systems. vol.~31, pp. 8580--8589 (2018)

\bibitem{Katz/etal/2017a}
Katz, G., Barrett, C.W., Dill, D.L., Julian, K., Kochenderfer, M.J.: Reluplex:
  An efficient {SMT} solver for verifying deep neural networks. In: {CAV} 2017.
  vol. 10426, pp. 97--117. Springer (2017)

\bibitem{Krizhevsky/etal/2012a}
Krizhevsky, A., Sutskever, I., Hinton, G.E.: Imagenet classification with deep
  convolutional neural networks. In: Advances in Neural Information Processing
  Systems 25 (2012)

\bibitem{Kurakin/etal/2017a}
Kurakin, A., Goodfellow, Ian J.and~Bengio, S.: Adversarial examples in the
  physical world. (ICLR)  \textbf{5} (february 2017)

\bibitem{Kwiatkowska/etal/2011}
Kwiatkowska, M., Norman, G., Parker, D.: Prism 4.0: Verification of
  probabilistic real-time systems. In: Computer Aided Verification. pp.
  585--591 (2011)

\bibitem{Langer/etal/2021a}
Langer, M., et~al.: What do we want from explainable artificial intelligence
  (xai)? a stakeholder perspective on xai and a conceptual model guiding
  interdisciplinary xai research. Artificial Intelligence  (Feb 2021)

\bibitem{Lauritzen/Spiegelhalter/88a}
Lauritzen, S.L., Spiegelhalter, D.J.: Local computations with probabilities on
  graphical structures and their application to expert systems. Journal of the
  Royal Statistical Society, Series B  \textbf{50}(2),  157--224 (1988)

\bibitem{Madry/etal/2019a}
Madry, A., Makelov, A., Schmidt, L., Tsipras, D., Vladu, A.: Towards deep
  learning models resistant to adversarial attacks (2019)

\bibitem{Mierswa/etal/2006a}
Mierswa, I., Wurst, M., Klinkenberg, R., Scholz, M., Euler, T.: {YALE}: {R}apid
  {P}rototyping for {C}omplex {D}ata {M}ining {T}asks. In: SIGKDD 2006. pp.
  935--940

\bibitem{Mitchell/etal/2019a}
Mitchell, M., et~al.: Model cards for model reporting. In: FAT* 2019. pp.
  220--229. Association for Computing Machinery (2019)

\bibitem{Morik/94a}
Morik, K.: Balanced Cooperative Modeling, vol.~4, pp. 295--318. Morgan Kaufmann
  (1994)

\bibitem{morik2021yes}
Morik, K., et~al.: Yes we care! -- certification for machine learning methods
  through the care label framework  (2021), arXiv preprint, arXiv:2105.10197

\bibitem{Murphy/2013a}
Murphy, K.P.: Machine Learning: a Probabilistic Perspective. MIT Press (2013)

\bibitem{Papernot/etal/2018a}
Papernot, N., et~al.: Technical report on the cleverhans v2.1.0 adversarial
  examples library (2018),
  \url{https://github.com/cleverhans-lab/cleverhans#readme}

\bibitem{Piatkowski/2018a}
Piatkowski, N.: Exponential Families on Resource-Constrained Systems. Ph.D.
  thesis, TU Dortmund University, Dortmund (2018),
  \url{https://eldorado.tu-dortmund.de/handle/2003/36877}

\bibitem{Piatkowski/etal/2016a}
Piatkowski, N., Lee, S., Morik, K.: Integer undirected graphical models for
  resource-constrained systems. Neurocomputing  \textbf{173}(1),  9--23
  (January 2016)

\bibitem{Piatkowski/etal/2014a}
Piatkowski, N., Sangkyun, L., Morik, K.: The integer approximation of
  undirected graphical models. In: {ICPRAM} 2014. pp. 296--304. SciTePress
  (2014)

\bibitem{Raji/etal/2020a}
Raji, I.D., et~al.: Closing the ai accountability gap: Defining an end-to-end
  framework for internal algorithmic auditing. In: FAT* 2020. pp. 33--44

\bibitem{Richards/etal/2020a}
Richards, J., Piorkowski, D., Hind, M., Houde, S., Mojsilovic, A.: A
  methodology for creating {AI} factsheets  (2020), arXiv preprint,
  arXiv:2006.13796

\bibitem{Ruan/etal/2018a}
Ruan, W., Huang, X., Kwiatkowska, M.: Reachability analysis of deep neural
  networkswith provable guarantees. IJCAI 2018

\bibitem{Samek/etal/2019a}
Samek, W., Montavon, G., Vedaldi, A., Hansen, L.K., M{\"u}ller, K.R. (eds.):
  Explainable AI: Interpreting, Explaining and Visualizing Deep Learning (2019)

\bibitem{Simonyan/Zisserman/2015a}
Simonyan, K., Zisserman, A.: Very deep convolutional networks for large-scale
  image recognition. In: {ICLR} 2015

\bibitem{Sokol/Flach/2020a}
Sokol, K., Flach, P.: Explainability fact sheets: a framework for systematic
  assessment of explainable approaches. In: {FAT*}. pp. 56--67. ACM (2020)

\bibitem{Strubell/etal/2020a}
Strubell, E., Ganesh, A., McCallum, A.: Energy and policy considerations for
  modern deep learning research. In: {AAAI} 2020. pp. 13693--13696 (2020)

\bibitem{Stutz/etal/2020a}
Stutz, D., Chandramoorthy, N., Hein, M., Schiele, B.: Bit error robustness for
  energy-efficient dnn accelerators. Proceedings of Machine Learning and
  Systems 3 pre-proceedings (MLSys 2021)  (2020), accepted at MLSys 2021

\bibitem{Szegedy/etal/2014a}
Szegedy, C., et~al.: Intriguing properties of neural networks. In:
  International Conference on Learning Representations (2014)

\bibitem{DBLP:conf/cvpr/TanCPVSHL19}
Tan, M., et~al.: Mnasnet: Platform-aware neural architecture search for mobile.
  In: CVPR 2019. pp. 2820--2828 (2019)

\bibitem{Tramer/Boneh/2019a}
Tram{\`{e}}r, F., Boneh, D.: Adversarial training and robustness for multiple
  perturbations. In: NeurIPS 2019. pp. 5858--5868 (2019)

\bibitem{Wu/etal/2020a}
Wu, M., Wicker, M., Ruan, W., Huang, X., Kwiatkowska, M.: A game-based
  approximate verification of deep neural networks with provable guarantees.
  Theor. Comput. Sci.  \textbf{807},  298--329 (2020)

\bibitem{Xiao/etal/2020a}
Xiao, L., Pennington, J., Schoenholz, S.: Disentangling trainability and
  generalization in deep neural networks. In: {ICML} 2020. vol.~119, pp.
  10462--10472

\end{thebibliography}

\clearpage

\end{document}